\documentclass[fleqn,10pt,twocolumn]{SWARM25}

\usepackage{subfigure}
\usepackage{graphicx} 
\usepackage[acronym,toc]{glossaries}
\usepackage{xcolor}
\usepackage{amsfonts}
\usepackage{amsmath}
\usepackage{amssymb}
\usepackage{subcaption}

\title{Understanding visual attention beehind bee-inspired UAV navigation}

\author{
Pranav Rajbhandari${}^{1\star}$,
Abhi Veda${}^{1\dagger \star}$,
Matthew Garratt${}^{1}$,
Mandyam Srinivasan${}^{2}$ and
Sridhar Ravi${}^{1}$
}
\speaker{Abhi Veda}
\affils{${}^{1}$ School of Engineering and Technology, University of New South Wales, Canberra, Australia,\\
${}^{2}$ Queensland Brain Institute, University of Queensland, Brisbane, Australia\\
${}^{\star}$ contributed equally to this paper.\\
(E-mail: 
prajbhan@cs.cmu.edu,
\{a.veda,
m.garratt,
sridhar.r\}@unsw.edu.au,
m.srinivasan@uq.edu.au)
}

\abstract{
Bio-inspired design is often used in autonomous UAV navigation due to the capacity of biological systems for flight and obstacle avoidance despite limited sensory and computational capabilities.
In particular, honeybees mainly use the sensory input of optic flow, the apparent motion of objects in their visual field, to navigate cluttered environments.
In our work, we train a Reinforcement Learning agent to navigate a tunnel with obstacles using only optic flow as sensory input.
We inspect the attention patterns of trained agents to determine the regions of optic flow on which they primarily base their motor decisions.
We find that agents trained in this way pay most attention to regions of discontinuity in optic flow, as well as regions with large optic flow magnitude.
The trained agents appear to navigate a cluttered tunnel by avoiding the obstacles that produce large optic flow, while maintaining a centered position in their environment, which resembles the behavior seen in flying insects.
This pattern persists across independently trained agents, which suggests that this could be a good strategy for developing a simple explicit control law for physical UAVs.
}


\begin{document}

\newcommand{\com}[1]{}
\newcommand{\todo}[1]{\textcolor{red}{TODO: #1}}



\newacronym{SHAP}{SHAP}{SHapley Additive exPlanations}
\newacronym{GCRL}{GCRL}{Goal Conditioned RL}
\newacronym{RL}{RL}{Reinforcement Learning}
\newacronym{MDP}{MDP}{Markov Decision Process}
\newacronym{POMDP}{POMDP}{Partially Observable MDP}
\newacronym{PPO}{PPO}{Proximal Policy Optimization}
\newacronym{CNN}{CNN}{Convolutional Neural Network}
\newacronym{UAV}{UAV}{Unmanned Aerial Vehicle}
\newacronym{MAV}{MAV}{Micro Aerial Vehicle}


\newcommand{\lrq}[1]{\lq#1\rq}

\newcommand{\meminisection}[1]{\textit{\textbf{#1}}}
\newcommand{\meminisubsection}[1]{\textit{#1}}


\newcommand{\imgwidth}[0]{0 pt}
\newcommand{\imgheight}[0]{0 pt}

\pagestyle{headings}

\maketitle


\section{Introduction}
\subsection{Bio-inspired autonomous navigation}
With the growing utilization of \gls{UAV} and \gls{MAV} systems for tasks like environment exploration/mapping, surveillance, and payload delivery, autonomous navigation of these systems has gained relevance.
These robots often have constraints, such as inability to use GPS and reduction in available sensors, due to their use cases or maximum payload.
However, the computational requirements of these systems remain large due to the high fidelity demands of autonomous aerial navigation.
To maintain performance under various constraints, navigation methods developed often take inspiration from biological organisms.
These methods are designed to be efficient and robust, as they mimic biological systems that have developed their capability through the process of evolution over millions of years.
They serve as an attractive alternative to traditional navigation methods, such as \cite{Lu2018,Chao2013,Chang2023,Gyagenda2022}.

In particular, the honeybee has been the subject of various studies for bio-inspired robotic navigation \cite{Srinivasan2010,Srinivasan2011}.
One of the main sensory inputs that bees utilize during flight is \textit{optic flow}, the relative motion of elements in the visual field of the observer \cite{Egelhaaf2025,Egelhaaf2023,beeopticflow}.
The bee's reliance on optic flow arises from limited sensory capabilities, including low visual acuity and resolution of honeybee eyes, limited neural capacity, and a lack of stereo vision.
Hence, it is unfeasible for them to rely solely on visual features for navigation. \cite{Currea2023,Jezeera2022}.
These constraints also make optic flow a useful sensory input for weight-constrained robotic control, as optic flow can be calculated from video captured by a single camera, which is relatively light compared to other sensory payloads.
This has led to the adaptation of honeybee capabilities for various robotic tasks, such as autonomous tunnel navigation \cite{Serres2017}, visual odometry \cite{Iida2000,Iida2003,RibeirodosSantos2023,Srinivasan2000}, landmark-based navigation \cite{Fry2005,Bianco2004,Cumbo2016} and autonomous landing \cite{Hammad2024,Goyal2023,Landing-fixed-wing-aircraft}.
These results benefit from the honeybee's capability for associative learning \cite{VonFrisch1974}, creating greater understanding of navigational principles of the organism through behavioral experiments.

While these works design systems whose control is inspired from honeybee behavior, our work uses a learning algorithm to train agents with sensory input similar to a bee to fly.
Our goal is to inspect the learned behavior to determine what the agent is paying most attention to in its sensory input.
Since there is evidence supporting the theory that bees use optic flow from different areas of vision for different control tasks \cite{Labhart1980,Lehrer1998,Giurfa1999,Seidl1981}, we investigate whether similar patterns arise from learning agents that use the same sensory information.

\subsection{Reinforcement Learning}
A \gls{MDP} is a framework useful for formalizing robotic optimization tasks.
\gls{MDP}s consist of a state space $S$, 
action space $A$, 
transition function ${\mathcal T}(S\mid S,A)$,
reward function $R\colon S\times A\times S\to \mathbb R$,
and discount factor $\gamma$.
An agent in an \gls{MDP} is described by its policy $\pi(A\mid S)$, which chooses a distribution over actions at each state.
Given an \gls{MDP} and an agent, we may generate a trajectory $(s_0,a_0,r_0),(s_1,a_1,r_1),\dots$, where $s_i\sim {\mathcal T}(\cdot\mid s_{i-1},a_{i-1})$, $a_i\sim \pi(\cdot\mid s_i)$, and $r_i=R(s_i,a_i,s_{i+1})$.
An agent's goal is to maximize the expected \textit{return} ${\mathbb E}[\sum_{i}\gamma^ir_i]$ over trajectories.
A \gls{POMDP} is an extension of this framework that includes 
an observation space $\Omega$
and an observation function $O(\Omega\mid S)$.
An agent in a \gls{POMDP} must choose action $a_i$ based on its \textit{belief}, a probability distribution over states.
We simplify this to a policy $\pi(A\mid \Omega)$ depending only on the current observation $o_i\sim O(\cdot \mid s_i)$.

\gls{RL} is a method to solve these tasks by repeatedly sampling trajectories to optimize an agent's policy.
Deep \gls{RL} uses deep neural networks to define the agent's policy (the \textit{policy network}), as well as for functions like the expected return from a state (the \textit{value network}).

In our work, we aim to train robotic agents to perform the task of avoiding obstacles while navigating a tunnel using bee-like optic flow observations.
To do this, we structure the task as a \gls{POMDP}, where agents are given reward for navigating successfully through the tunnel, and a penalty for crashing.
Our goal is to inspect the policy of an \gls{RL} agent that learned this task, to find out to which portion of each observation the agent pays most attention.

\subsection{Explainable AI/Shapley Values}
To understand the attention patterns of a trained agent, we use methods from Explainable AI, a field concerned with exactly this problem.
Specifically, \gls{SHAP} is a method rooted in game theory that assigns values to a set of input features that represent how much each feature contributed to a decision \cite{shap_value}.
In our case, the features will be specific regions of observed optic flow.
We report the absolute \gls{SHAP} value since we care about the magnitude of the contribution of each feature, as opposed to whether it had a positive or negative effect.
We use the attention patterns of trained neural networks to hypothesize how real bees may utilize different regions of optic flow for a particular task.

\section{Previous Work}

\subsection{Buggy robotics}
Insects, albeit limited in computational capacity ($\approx 10^4$ times less neurons in the brain than humans \cite{Makarova2021}), solve similar navigation problems as \gls{MAV}s with remarkable performance, even with a miniature form factor and no complex sensory system.
This has resulted in insect-inspired systems for autonomous navigation tasks \cite{Strydom2016,deCroon2022,Zeil2022,Hu2025,Duan2014}.
Roboticists mimic insects both via physical robot hardware that imitates morphological design and vision capabilities \cite{Ratti2012,Budholiya2021,Hui2019,Geronel2025}, as well as via robot behavior that imitates insect-based navigational strategies \cite{Aljalaud2023,Lehnert2019,Lei2024,Denuelle2015,Sabo2016}.
These low-resource vision and computation capabilities of insects have also influenced neuromorphic systems and techniques for navigation \cite{Schoepe2024,Paredes-Valls2024}.
These methods allow for rapid visual information to be utilized to drive navigational systems at lower power and computational cost compared to  traditional hardware (such as SLAM, proximity sensors, or depth cameras).

\subsection{Explainable \gls{UAV} navigation}
Previous studies have developed sophisticated \gls{UAV} navigation methods utilizing sensory input from LiDAR, depth sensors, RGB cameras, or a combination of these \cite{Ren2025,Dissanayaka2024,He2021}.
These works analyze the learned navigation behavior through the use of \gls{SHAP} values or Grad-CAM \cite{Selvaraju2017}.
This allows for an interpretation of how the input features are utilized, giving insight into the working of the control algorithms.
However, none of these methods are directly bio-inspired, and none utilize optical flow for sensory input.
Through our work, we aim to address this gap by utilizing optic flow in an autonomous navigation system, and exploring how an autonomous system interprets this sensory input.
This method of explaining a bio-inspired autonomous system stems from ideas presented in \cite{opposite_day,Iida2016BiologicallyRobotics}.
We use our results, which reflect how an autonomous agent formulates a control behavior from optic flow, to hypothesize how real bees navigate based on the same sensory input.

\section{Simulation Environment}
\begin{figure}
    \begin{center}
        \includegraphics[width=.69\linewidth]{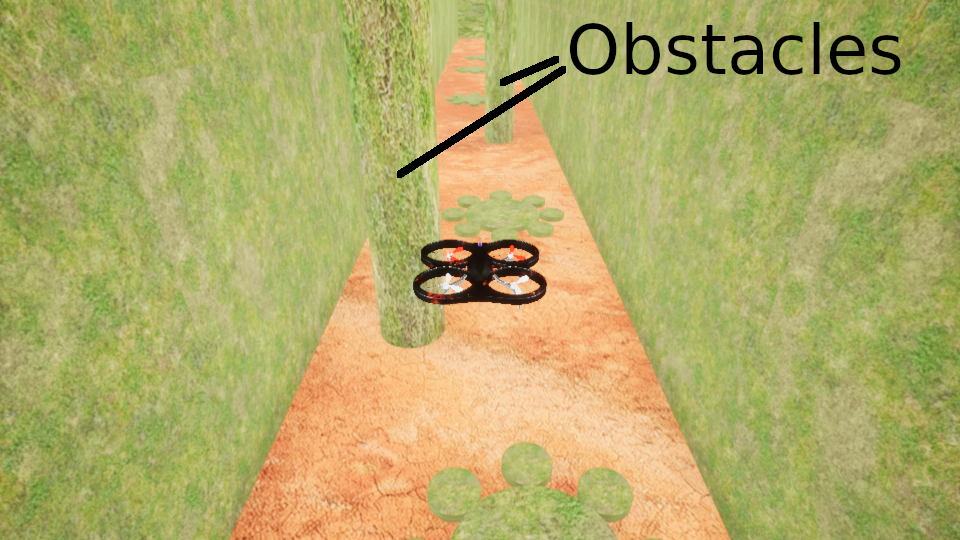}
        \caption{Screenshot of the AirSim simulation environment}
        \label{fig:env-img}
    \end{center}
\end{figure}

We conduct our experiments in a simulated UAV environment, using the AirSim platform \cite{airsim} for the Unreal physics simulation engine \cite{unrealengine}.
The agent is a quadrotor drone that is traveling through a constrained tunnel containing cylindrical obstacles (displayed in Fig.~\ref{fig:env-img}).
There are six tunnels, each with different widths and obstacle configurations.
The agent's goal is to reach the end of the tunnel without colliding with the walls or obstacles.
We structure this as a \gls{POMDP} with the OpenAI Gymnasium environment\footnote{Our experiment implementation is available at \cite{beehavior_code}.} \cite{openaigym}.
We define elements of the state space $S$ as states of the physics simulation, and we define sampling from the transition function $\mathcal T$ as running the physics simulation for a duration $\Delta t=0.05~s$.
We set discount factor $\gamma=0.99$.

\subsection{Observation Space}

To imitate sensory input of real bees during flight, we allow the agent to observe optic flow at each timestep from a camera with field of view $120^\circ$ mounted to the front of the drone.
The optic flow is calculated geometrically, using knowledge of the drone's true velocity and distance from each obstacle.
In the case of a real vehicle, there exist algorithms to calculate optic flow from single camera input \cite{Farnebck2003,Zhai2021,Srinivasan1994AnEgomotion,Benosman2012}.
We calculate optic flow from the formulation presented in \cite{Li2016MetricMethod}:
\begin{align}
\setlength\arraycolsep{2pt} 
\begin{bmatrix}
\dot{u} \\
\dot{v}
\end{bmatrix}
=
\begin{bmatrix}
\frac{f_x}{d_{u,v}} & 0 & -\frac{u}{d_{u,v}} & -\frac{uv}{f_x} & -\frac{f_x^2 + u^2}{f_x} & -v \\
0 & \frac{f_y}{d_{u,v}} & -\frac{v}{d_{u,v}} & -\frac{f_y^2 + u^2}{f_y} & \frac{uv}{f_y} & u
\end{bmatrix}
\begin{bmatrix}
V \\
W
\end{bmatrix}
\end{align}

Where $f_x$ and $f_y$ are the focal lengths of the camera in $x$ and $y$ direction,
$u$ and $v$ are the horizontal and vertical pixel coordinates,
$V=[\dot{x}, \dot{y}, \dot{z}]^T$ is the linear velocity of the vehicle in cartesian coordinates,
$W=[W_x,W_y,W_z]^T$ is the angular velocity,
and $d_{u,v}$ is the distance of the vehicle from the object at pixel $(u,v)$.

Explicitly, an element of the observation space $\Omega$ is an array containing the optic flow magnitude and orientation at each pixel in the camera's field of view.
We represent this as an \lrq{image} with three channels: 
the optic flow magnitude,
and the horizontal and vertical components of the normalized optic flow vector.
We set angular velocity to zero during optic flow calculation ($W=\mathbf{0}$), since we found that the contribution from the drone's angular velocity is often much larger than the contribution from its linear motion.\footnote[3]{This can be simulated on physical systems either through measuring angular velocity, or through estimating it from observed optic flow.}
We visualize an observation in Fig. \ref{fig:pay-attention}(b), where the agent is traveling forward and right.
Here, we display (unnormalized) optic flow vectors as black arrows over a heatmap of the optic flow magnitude.

\subsection{Action Space}
The agent controls the acceleration of the drone in the basis where the $+x$ and $+y$ directions correspond to the drone's forward and starboard directions respectively.
We project the $+x$ and $+y$ direction onto the global $xy$ plane, so that both of these are normal to the global $z$ axis.
The acceleration is added to a target velocity, which is then sent as a velocity command to the AirSim platform.
Both the acceleration and the target velocity have a limited magnitude on each dimension, which we denote by $a_{\text{max}}=3~m/s^2$ and $v_{\text{max}}=2~m/s$ respectively.
We represent the action space $A$ as $\{\textbf{a}\in {\mathbb R}^2:||\textbf{a}||_\infty\leq a_{\text{max}}\}$.
We fix the drone's height above the ground in our experiments, as our environment contains only column obstacles and \lrq{flowers} below the drone's flying height.

\subsection{Reward Function/Termination}

Let $m_i$ be the maximum global $x$ coordinate of the agent throughout timesteps $[0,\dots,i]$.
We set the reward $r_i:=m_{i+1}-m_i$.
The reward is positive if the agent increases the furthest it has traveled, otherwise it is zero.
The episode terminates with a penalty $(r_i=-1)$ if the agent crashes at timestep $i$.
The episode terminates with no penalty if the agent successfully navigates the tunnel or runs out of time (after $30~s$).
The motivation behind our reward function is to encourage forward travel while not directly penalizing a path that slows to avoid obstacles.

\subsection{Agent Policy}
\begin{figure}[ht!]
    \begin{center}
        \includegraphics[width=\linewidth]{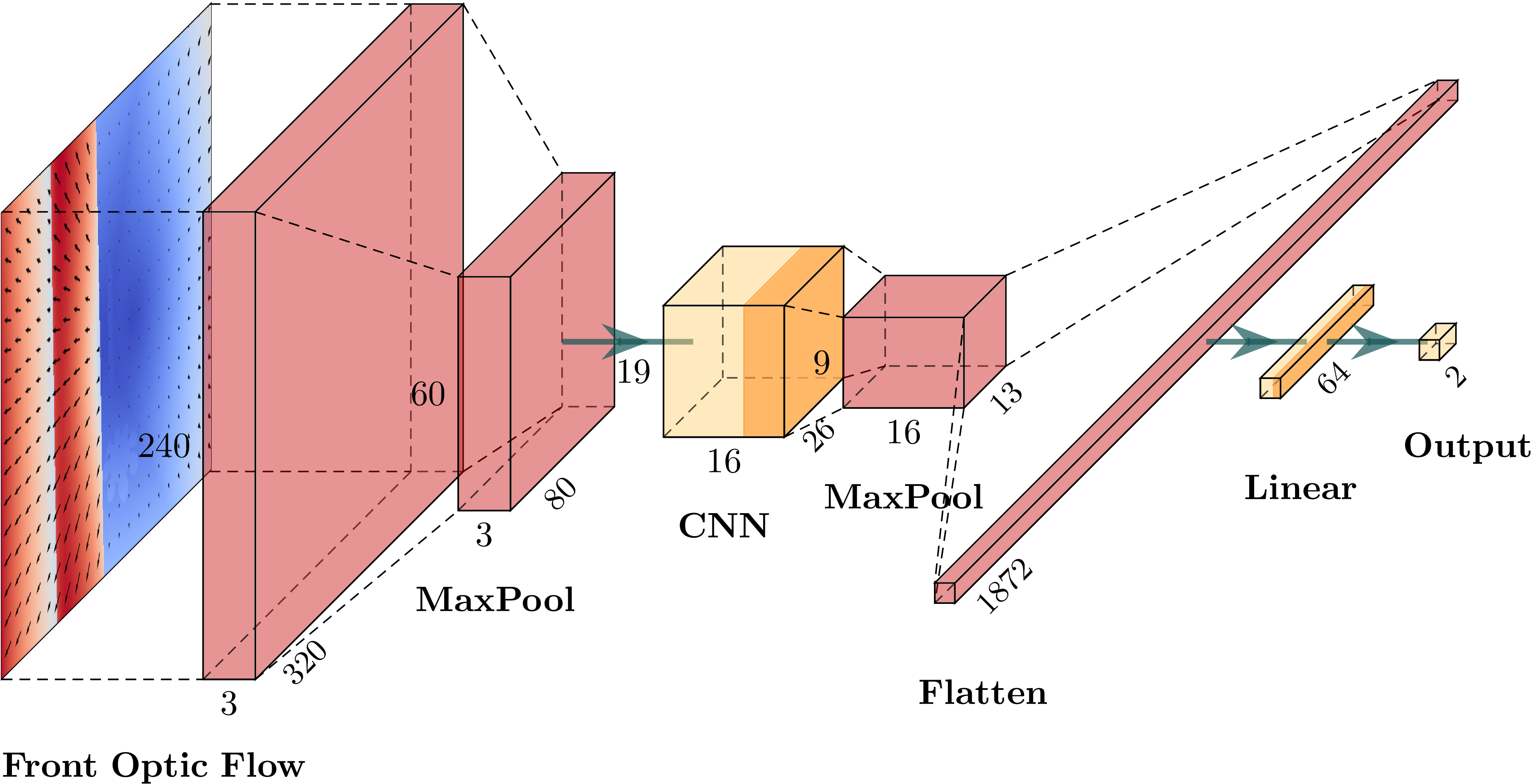}
        \caption{Policy Network Architecture.
        Yellow boxes represent transitions with learnable weights.}
        \label{fig:nn-shape}
    \end{center}
\end{figure}

To train an agent in this environment we use \gls{PPO}, a Deep \gls{RL} algorithm that defines policy and value networks \cite{ppo}.
For the policy network, we use a network architecture whose layers with learnable weights are a \gls{CNN} layer with ReLU activation, a fully connected linear layer with Tanh activation, and a linear output layer (Fig. \ref{fig:nn-shape}).
The architecture of the value network is identical to that of the policy network, with the exception of the output layer, which maps to a one-dimensional output instead.
We chose this architecture since it is relatively simple and performed well in initial experiments.


\section{Results}

We train our agent in six different tunnel configurations, which vary in width and in number of obstacles.
At the start of each trial, the agent's position is randomized at the beginning of a tunnel randomly chosen from the six.
We train for 100 epochs, with each epoch consisting of 1024 timesteps ($\approx50~s$ of simulation).

We train four agents independently in this way, so that we may inspect general trends in attention patterns rather than the learned attention of any one specific trained agent.
We sample a few trajectories from one of these trained agents navigating a tunnel.
We use the recorded optic flows at each timestep in these trajectories to obtain \gls{SHAP} values for each of the four agents on these observations.
We take the absolute value of these \gls{SHAP} values, apply Gaussian smoothing, and take the average across all four agents.
The resulting attention patterns for a single timestep are displayed in Figures \ref{fig:pay-attention} and \ref{fig:model-compare}.

\renewcommand{\imgwidth}{.31\linewidth}
\begin{figure*}[ht!]
\centering
\subfigure[ Visual scene]{
\includegraphics[width=.27\linewidth]{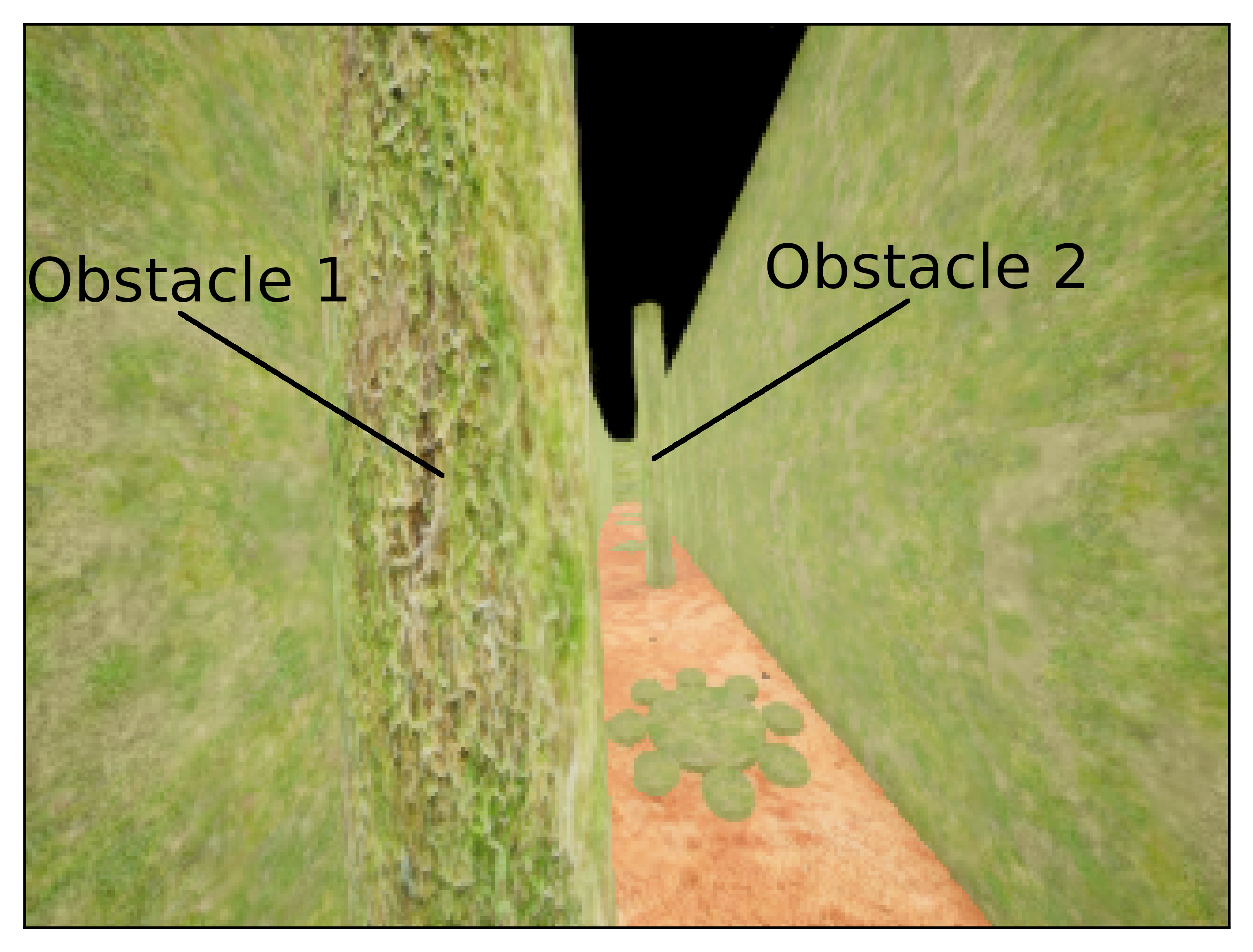}
}
\subfigure[ Optic flow perceived by agent]{
\includegraphics[width=\imgwidth]{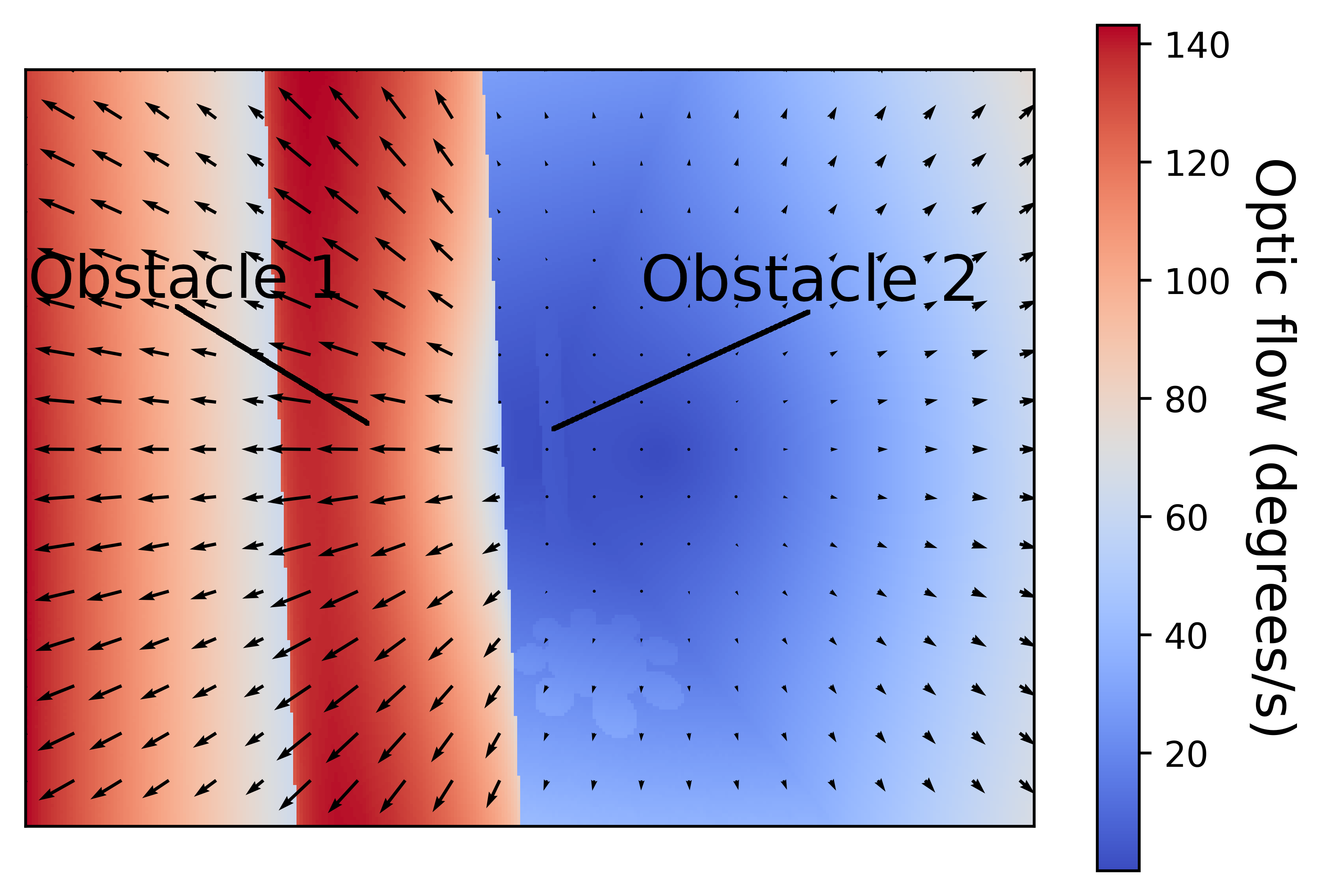}
}
\subfigure[ Average sum of absolute \gls{SHAP} values]{
\includegraphics[width=\imgwidth]{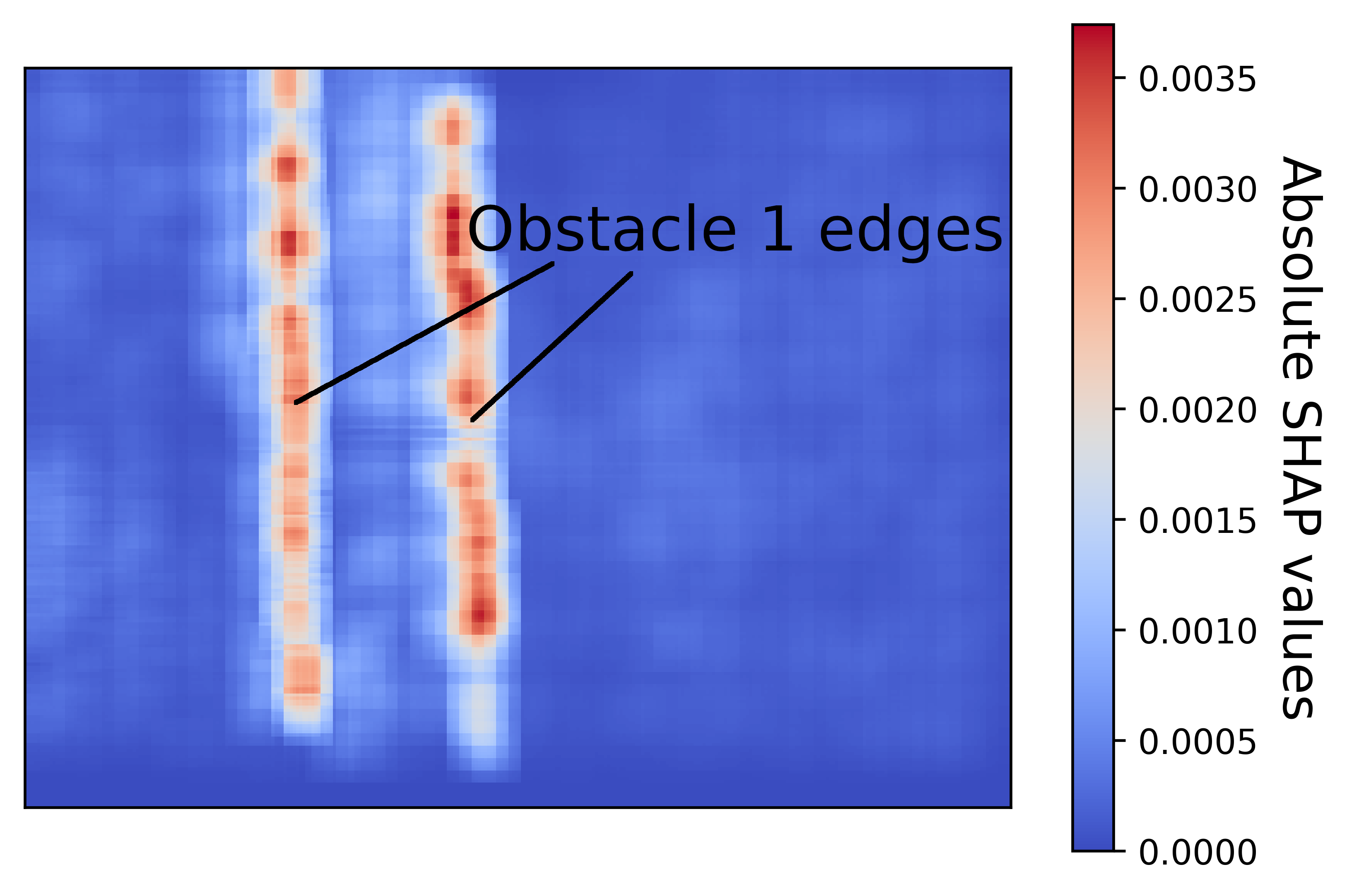}
}
\caption{The attention pattern of trained agents flying through a tunnel, averaged over four independently trained agents.}
\label{fig:pay-attention}
\end{figure*}

We find that the trained agents are successful in navigating the tunnels using optic flow observations.
Their trajectories (displayed in Fig. \ref{fig:bee-flying}) appear to center the agent in the non-obstructed region of the tunnel.


\renewcommand{\imgwidth}{.4\linewidth}
\begin{figure*}[htb!]
\centering
\subfigure[ Easier tunnel]{
\includegraphics[width=\imgwidth]{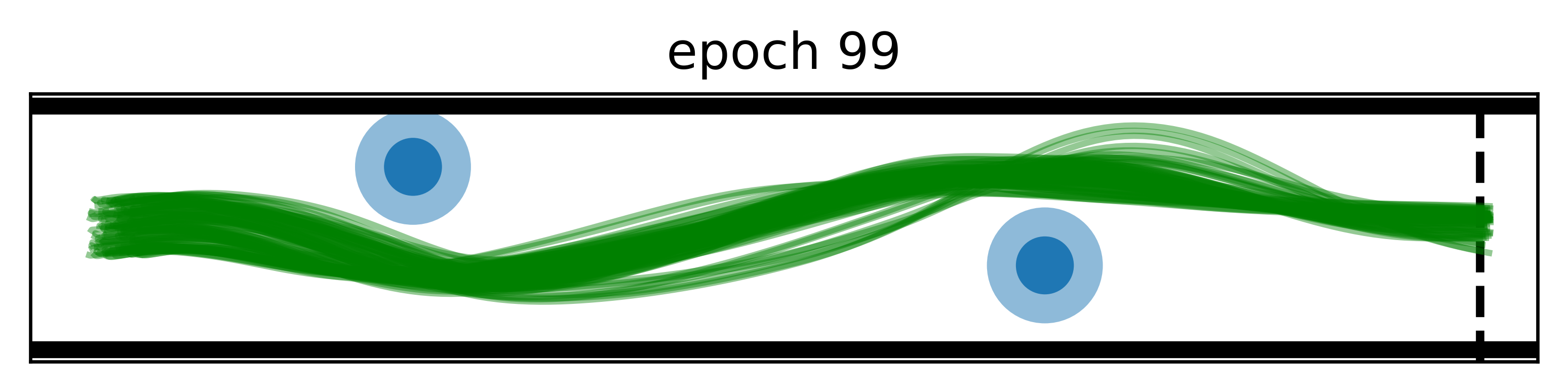}
}
\subfigure[ Difficult tunnel]{
\includegraphics[width=\imgwidth]{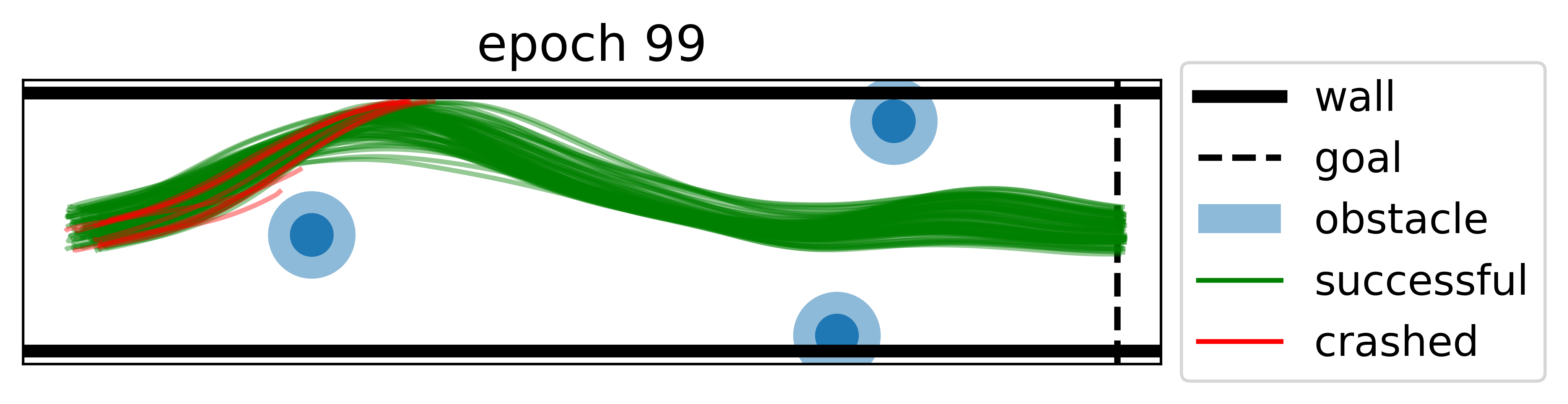}
}
\subfigure[ Success rate in easier tunnel throughout training]{
\includegraphics[width=\imgwidth]{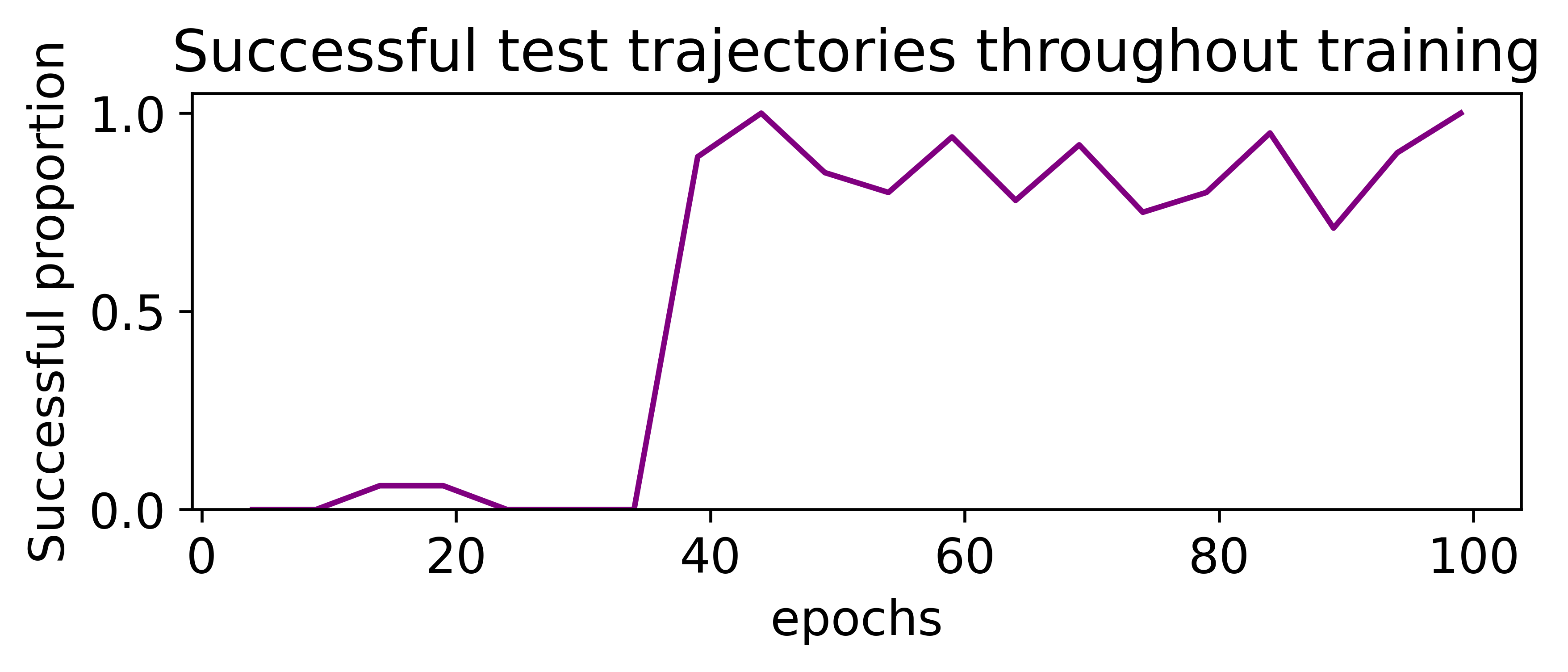}
}
\subfigure[ Success rate in difficult tunnel throughout training]{
\includegraphics[width=\imgwidth]{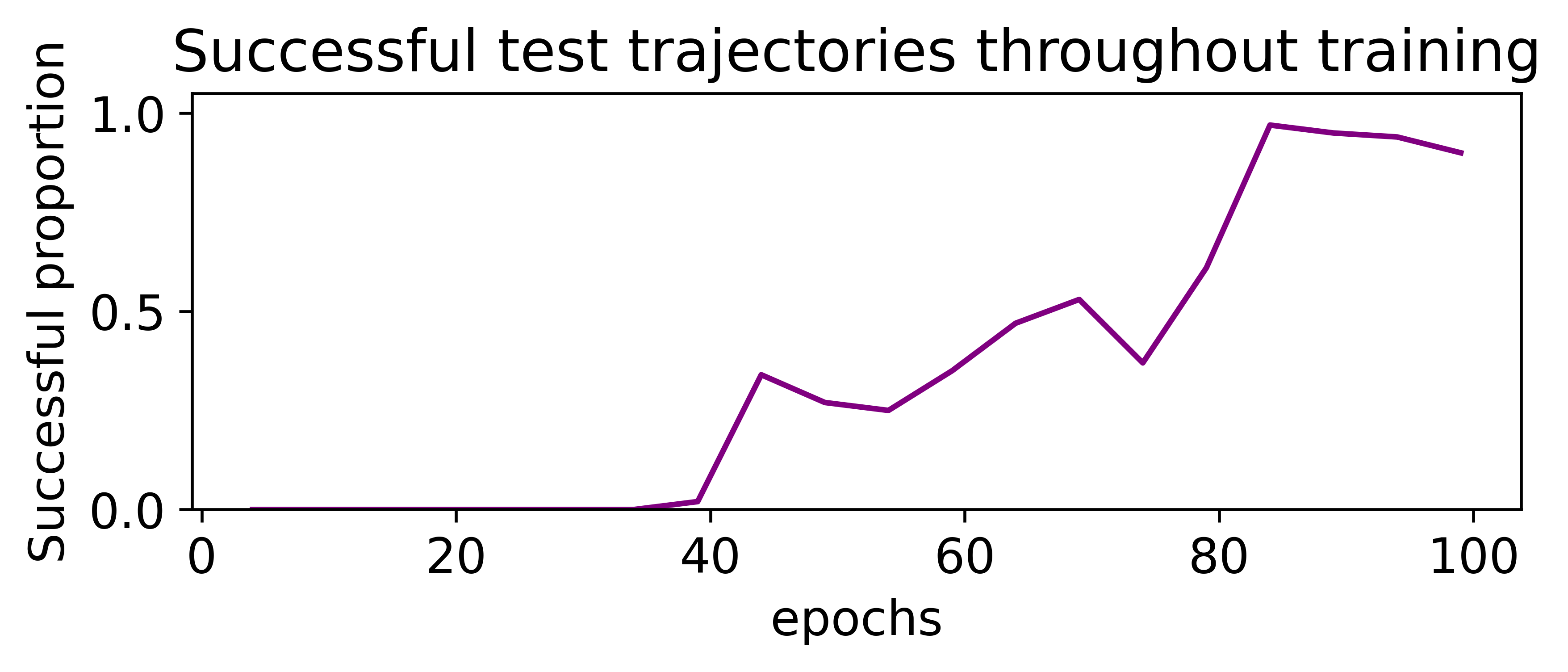}
}
\caption{Trajectories from a trained agent flying through different tunnels: (a) Easier tunnel with two obstacles placed off-center, (b) Difficult tunnel with three obstacles, one placed centrally and two placed off-center.
}
\label{fig:bee-flying}
\end{figure*}

We find that regions of discontinuity in optic flow magnitude make a disproportionally larger contribution to the agent's decisions.
This is seen in Fig. \ref{fig:pay-attention}, where the agents pay most attention to the nearest edges of an obstacle.
Aside from this, the agents seem to pay attention to regions of high optic flow, such as the center of the nearest obstacle and the closest parts of the tunnel walls in Fig. \ref{fig:pay-attention}.
We observe this attention pattern throughout the trajectories, in addition to the example shown.

These trajectories and attention patterns could be explained by the learned behavior being to move away from regions with high optic flow, while placing the most value on regions with discontinuities in the optic flow.
The reason for the larger attention towards discontinuities could be that discontinuities are a good visual indicator of an obstacle, which create large optic flow values in a small portion of the visual field.
If the agent did not handle these cases specially, the thinness of the obstacle may outbalance the magnitude of the optic flow values, resulting in not avoiding a close obstacle due to its small size.

\renewcommand{\imgwidth}{\linewidth}
\begin{figure*}[ht!]
\centering
\includegraphics[width=\imgwidth]{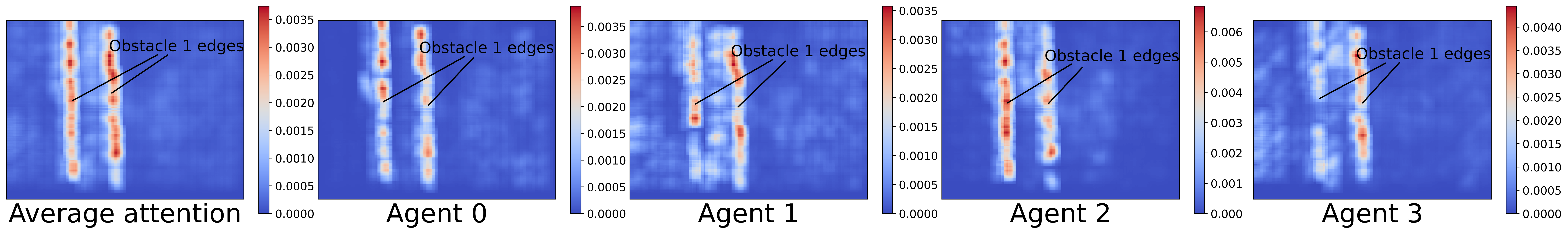}
\caption{Comparison of attention patterns for four independently trained agents.
The average attention is largest at nearby obstacle edges; this pattern exists in each of the four agent policy networks as well.
}
\label{fig:model-compare}
\end{figure*}

These attention patterns seem to be learned independently by all four of our agents (Fig. \ref{fig:model-compare}), which indicates this arises as a result of the task and observation space as opposed to arising from a quirk of a particular agent policy network.

\section{Discussion}

In our experiments, our trained agents appear to balance the optic flow perceived in the left and right regions of their visual field to maintain a centered flight path.
We find that the agents pay most attention to nearby edges of obstacles, where discontinuities in optic flow are most significant (Fig. \ref{fig:model-compare}).
This results in the the well documented centering behavior of honeybees \cite{Serres2008AWall-following,Baird2005VisualHoneybees} appearing in the agents' trajectories (Fig. \ref{fig:bee-flying}), as well as detection and obstacle avoidance evident from lateral deviation and reduction in forward speed.
The similarities of the trajectories to navigation patterns found in bees \cite{Serres2017,Srinivasan2011} indicate that real bees also respond strongly to discontinuities in optic flow, such as those induced by the edges of obstacles.
This also suggests that an effective computational navigation strategy should involve discontinuity detection, as opposed to regulating raw optic flow values.

\section{Conclusion and Future Work}

We demonstrate that behavior learned through deep \gls{RL} with the same sensory input as a bee induces similar navigation capability as the organism.
The underlying attention pattern for this learned behavior is a novel contribution and can inspire a formalized control scheme for an autonomous \gls{UAV}.
In future work, we plan to use similar methods in a  \acrlong{GCRL} environment, and compare the trained agent's attention patterns for various goals such as moving forward, landing, or passing through gaps.
We also plan to define control laws inspired by the demonstrated agent behavior that induce similar \gls{SHAP} attention patterns to the trained agents.



\bibliographystyle{unsrt} 
\bibliography{refs}

\end{document}